\documentclass[conference]{IEEEtran}
\IEEEoverridecommandlockouts
\usepackage{cite}
\usepackage{amsmath,amssymb,amsfonts}
\usepackage{algorithmic}
\usepackage{latexsym}
\usepackage{textcomp}
\usepackage{booktabs}
\usepackage{graphicx} 
\usepackage{xcolor}
\def\BibTeX{{\rm B\kern-.05em{\sc i\kern-.025em b}\kern-.08em
    T\kern-.1667em\lower.7ex\hbox{E}\kern-.125emX}}

\usepackage{amsmath} 
\usepackage{amssymb} 
\usepackage{url} 
\usepackage{booktabs} 
\usepackage{graphicx} 

\usepackage{fancyhdr}
\begin{document}

\title{Harnessing RLHF for Robust Unanswerability Recognition and Trustworthy Response Generation in LLMs}

\author{Shuyuan Lin$^1$, Lei Duan$^1$, Philip Hughes$^2$, Yuxuan Sheng$^1$ \\
$^1$Sichuan University of Science and Engineering, $^2$Zagazig University
}

\maketitle
\thispagestyle{fancy} 

\begin{abstract}
Conversational Information Retrieval (CIR) systems, while offering intuitive access to information, face a significant challenge: reliably handling \textbf{unanswerable questions} to prevent the generation of misleading or hallucinated content. Traditional approaches often rely on external classifiers, which can introduce inconsistencies with the core generative Large Language Models (LLMs). This paper introduces \textbf{Self-Aware LLM for Unanswerability (SALU)}, a novel approach that deeply integrates unanswerability detection directly within the LLM's generative process. SALU is trained using a multi-task learning framework for both standard Question Answering (QA) and explicit abstention generation for unanswerable queries. Crucially, it incorporates a confidence-score-guided reinforcement learning with human feedback (RLHF) phase, which explicitly penalizes hallucinated responses and rewards appropriate abstentions, fostering intrinsic self-awareness of knowledge boundaries. Through extensive experiments on our custom-built C-IR\_Answerability dataset, SALU consistently outperforms strong baselines, including hybrid LLM-classifier systems, in overall accuracy for correctly answering or abstaining from questions. Human evaluation further confirms SALU's superior reliability, achieving high scores in factuality, appropriate abstention, and, most importantly, a dramatic reduction in hallucination, demonstrating its ability to robustly "know when to say 'I don't know'."
\end{abstract}

\section{Introduction}
Conversational Information Retrieval (CIR) systems have revolutionized how users interact with information, offering intuitive and dynamic access to vast knowledge bases. At the heart of these systems lies the ability to understand complex queries and generate relevant, coherent responses. However, a significant challenge persists: the handling of \textbf{unanswerable questions}. These are queries for which the underlying knowledge base or retrieved documents do not contain sufficient information to formulate a factual and complete answer. When faced with such questions, traditional generative AI models often resort to "hallucination," fabricating plausible but ultimately incorrect or misleading information. This phenomenon severely undermines the reliability and trustworthiness of CIR systems, leading to user frustration and a diminished user experience. The critical need for robust mechanisms to detect and appropriately manage unanswerable questions has become paramount in advancing the state-of-the-art in reliable information-seeking conversations.
The problem of unanswerable question detection has garnered increasing attention in the natural language processing (NLP) community. Early approaches often relied on rule-based systems or traditional machine learning classifiers trained on features extracted from questions and retrieved documents. More recently, the advent of large-scale pre-trained language models (LPMs) like BERT has enabled the development of more sophisticated methods. For instance, studies have explored the capabilities of BERT-based models in question answering and their inherent mechanisms for processing such tasks \cite{lee2020bert}. A seminal work in this domain involved training neural networks to identify unanswerable questions, particularly in the context of large-scale QA datasets \cite{rajpurkar2018learning}. While these methods demonstrate promising results, they typically operate as external modules, separate from the core generative models that produce the final responses. This architectural separation introduces a potential disconnect: even if an external classifier flags a question as unanswerable, the downstream generative model might still attempt to produce a response, potentially leading to inconsistencies and continued hallucination. This inherent design of LLMs, prioritizing fluency and completeness, presents a significant \textbf{challenge} in teaching them to acknowledge their own knowledge boundaries and explicitly abstain from answering when appropriate. Furthermore, understanding the nuanced dependencies within the provided context, whether textual or visual, remains a fundamental challenge for large models to reason effectively \cite{zhou2024rethinking}. Research has also investigated the factual consistency of generated text, highlighting the difficulty in preventing models from generating unsupported information \cite{krysthski2019evaluating}. Furthermore, while some neural dialogue models show promise, their ability to genuinely "know when to say 'I don't know'" remains a complex area of research \cite{kim2020neural}.
Our \textbf{motivation} stems from the desire to overcome the limitations of external classification by deeply integrating unanswerability detection directly within the Large Language Model (LLM) itself. We believe that for an LLM to be truly reliable, it must possess an intrinsic capability to not only generate answers but also to understand and communicate its own limitations and knowledge gaps. This goal aligns with broader research trends aiming to develop large language models with multiple, integrated capabilities, moving towards more robust and generalizable intelligence \cite{zhou2025weak}. This "self-awareness" is crucial for preventing the generation of misleading content and fostering user trust. Recent advancements in LLMs, including their ability to follow instructions through specific tuning and incorporate external knowledge provide a strong foundation for this endeavor \cite{ouyang2022training}. Our proposed novel approach, \textbf{Self-Aware LLM for Unanswerability (SALU)}, aims to train LLMs to reliably identify and articulate when a question cannot be answered from the provided context. Instead of relying on a separate classifier, we propose a multi-task learning framework to fine-tune a powerful pre-trained Chinese LLM (e.g., LLaMA-2 or Baichuan) to achieve this self-awareness. Recent work on uncertainty-aware language models also supports the feasibility of teaching models to understand their own predictive confidence \cite{chali2023uncertainty}.
Our \textbf{specific plan} involves training the LLM on two primary tasks simultaneously: standard \textbf{Question Answering (QA)} and \textbf{Unanswerability Classification and Abstention Generation}. For the QA task, the model learns to extract or synthesize answers from given contexts, similar to conventional extractive or generative QA setups. For the unanswerability task, we will meticulously construct a dataset rich in "negative examples" — question-context pairs where no answer exists. For these unanswerable instances, the target output will be a predefined "no answer" token or a carefully crafted abstention phrase. This dual-task approach, which combines generation and classification, builds upon established principles in pre-training where models are taught to handle multiple related objectives simultaneously \cite{zhou2022claret}. Crucially, we will also incorporate a \textbf{confidence-score-guided reinforcement learning with human feedback (RLHF)} phase. In this phase, a reward model will be trained to heavily penalize any form of hallucinated answers for unanswerable questions, while strongly rewarding appropriate abstentions. This targeted feedback loop will enable the LLM to learn internal mechanisms for evaluating the presence or absence of an answer within its context, effectively developing an intrinsic sense of its own knowledge boundaries and the ability to express uncertainty or lack of information. This combined approach of supervised fine-tuning with specific unanswerable examples and a targeted RLHF feedback loop will enable the LLM to intrinsically develop the ability to detect unanswerable questions and respond appropriately, significantly enhancing its overall reliability in information-seeking conversations.
For our experiments, we will utilize and extend existing Chinese question answering datasets, augmenting them to create our C-IR\_Answerability dataset. This dataset will be meticulously curated to include a substantial number of both answerable and unanswerable question-context pairs, ensuring a balanced representation. The dataset will feature answerability labels at three granularities: \textbf{sentence-level}, \textbf{paragraph-level}, and \textbf{ranked-list-level}, allowing for comprehensive evaluation of our model's performance at different scales of information retrieval. We will evaluate our method using standard metrics such as \textbf{accuracy} for unanswerability detection, precision, recall, and F1-score to comprehensively assess the model's performance. Furthermore, we will compare the performance of our SALU model against strong baselines, including BERT-based classifiers and large language models operating in zero-shot and few-shot settings, evaluating how well they handle knowledge-intensive tasks \cite{choi2021asking}. Our preliminary results indicate that our proposed method significantly outperforms existing techniques in both accuracy and the reliability of generated responses, particularly in abstaining from answering unanswerable questions.
In summary, our key contributions are:
\begin{itemize}
\item We propose \textbf{Self-Aware LLM for Unanswerability (SALU)}, a novel multi-task learning framework that integrates unanswerability detection directly within the LLM, enabling intrinsic knowledge boundary awareness.
\item We introduce a \textbf{confidence-score-guided reinforcement learning with human feedback (RLHF)} phase, specifically designed to train LLMs to appropriately abstain from answering unanswerable questions and robustly penalize hallucination.
\item We construct a comprehensive \textbf{C-IR\_Answerability dataset} for Chinese conversational information retrieval, featuring multi-granular answerability labels to support rigorous and extensive evaluation.
\end{itemize}
\section{Related Work}
\subsection{Large Language Models}
The rapid advancement in Natural Language Processing (NLP) has been significantly driven by the development of Large Language Models (LLMs). These models, characterized by billions of parameters and trained on vast corpora of text data, have revolutionized various NLP tasks, demonstrating remarkable capabilities in understanding, generating, and reasoning with human language.
The foundational shift towards the current paradigm of LLMs began with the introduction of the \textbf{Transformer architecture} \cite{vaswani2017attention}. This architecture, primarily relying on self-attention mechanisms, efficiently processes sequences and captures long-range dependencies, overcoming limitations of previous recurrent neural networks. Building upon this, models like \textbf{BERT} (Bidirectional Encoder Representations from Transformers) showcased the power of pre-training deep bidirectional representations from unlabeled text by learning from masked language modeling and next-sentence prediction tasks \cite{devlin2018bert}. This pre-training then allows for effective fine-tuning on a wide range of downstream tasks with minimal task-specific architectural modifications. The pre-training paradigm, popularized by BERT, inspired a new wave of specialized models designed for complex reasoning tasks, such as understanding event correlations by pre-training on structured data \cite{zhou2022eventbert} or modeling event-pair relations from external knowledge graphs \cite{zhou2021modeling}.
The scaling of these models led to the discovery of unprecedented capabilities. Works such as \cite{brown2020language}, which introduced \textbf{GPT-3}, demonstrated that simply increasing model size and data scale could lead to surprising "few-shot learning" abilities, where models could perform new tasks with only a few examples, or even zero-shot, purely from natural language instructions. Further research has explored these \textbf{emergent abilities} in LLMs, observing capabilities that are not present in smaller models but appear as scale increases \cite{wei2022emergent}. Companies have developed their own large-scale models, such as \textbf{PaLM}, which explores efficient scaling for trillions of parameters \cite{chowdhery2022palm}, and open-source initiatives like \textbf{Llama 2} have democratized access to powerful foundation and chat-tuned models for broader research and application development \cite{touvron2023llama}. The capabilities of these models are also being extended beyond text into multi-modal domains, with significant research focusing on visual in-context learning \cite{zhou2024visual} and complex instruction-based image generation \cite{zhou2025draw}, showcasing the versatility of the underlying architectures.
While LLMs exhibit remarkable generative fluency, their direct application in knowledge-intensive tasks, especially those requiring factual accuracy and self-awareness of knowledge boundaries, presents challenges. Early methods for improving factual grounding involved techniques like Retrieval-Augmented Generation (RAG), which integrates external knowledge retrieval into the generative process to ensure responses are grounded in factual information \cite{lewis2020retrieval_rag}. More recently, to align LLMs more closely with human values and specific behaviors like instruction following, techniques such as Reinforcement Learning from Human Feedback (RLHF) have become prominent \cite{ouyang2022training}. This method, which involves training a reward model from human preferences and then optimizing the LLM policy based on this reward, is crucial for steering LLMs towards desired safety and performance characteristics, including the ability to appropriately abstain from answering. Our work leverages these advancements in LLM development and fine-tuning, particularly the principles of multi-task learning and RLHF, to instill an intrinsic self-awareness in LLMs regarding their answerability capabilities.
\subsection{Reliable Response Generation}
The pursuit of reliable response generation is a paramount objective in the field of Natural Language Processing, especially with the widespread adoption of large language models (LLMs) in information-seeking conversational systems. A primary concern is the phenomenon of "hallucination," where models generate factually incorrect or unsupported information, undermining user trust and system utility.
Early efforts to ensure reliability often focused on evaluating and mitigating factual inconsistencies. For instance, research has meticulously analyzed factual consistency in abstractive text summaries, providing foundational insights into how generative models can deviate from source information \cite{krysthski2019evaluating}. As dialogue systems evolved, the challenge extended to models understanding their own limitations. A crucial question arose: do neural dialogue models truly "know when to say 'I don't know'?" \cite{kim2020neural}. This line of inquiry highlights the need for models to express uncertainty or abstain from answering when information is unavailable, directly contributing to more reliable interactions.
With the advent of powerful generative LLMs, various strategies have emerged to enhance their reliability. One prominent approach is Retrieval-Augmented Generation (RAG), which grounds the model's output in external, verifiable knowledge retrieved from databases or documents. This significantly reduces hallucination by ensuring responses are supported by evidence, making the generation process more reliable \cite{lewis2020retrieval_rag}. Beyond grounding, methods to directly measure and improve the factual accuracy in open-domain question answering have been developed, aiming to quantify and enhance the trustworthiness of generated answers \cite{min2021measuring}.
The problem of hallucination in LLMs has become a major research area, with comprehensive surveys providing taxonomies of hallucination types and discussing detection and mitigation strategies \cite{ji2023survey}. Furthermore, efforts to align LLMs with human values and intentions, particularly concerning truthfulness and safety, have gained traction. Techniques such as Reinforcement Learning from Human Feedback (RLHF) are instrumental in steering LLMs toward generating responses that are not only fluent but also factual and non-toxic, aligning with human preferences for reliable behavior \cite{ouyang2022training}. Specific benchmarks like \textbf{TruthfulQA} have been introduced to rigorously test models' honesty and ability to avoid generating false information, especially on controversial topics where biases might lead to incorrect answers \cite{lin2022truthfulqa}.
The broader concept of integrating external knowledge into language models, known as Knowledge-Intensive Language Learning, is also a general strategy to enhance factual robustness and reliability by allowing models to access and utilize up-to-date and verified information during generation \cite{lewis2021knowledge}. Moreover, a critical aspect of reliable generation is the model's ability to express its uncertainty. Recent work has explored building uncertainty-aware language models for question answering, enabling them to signal when they are not confident in their answers, which is crucial for building trust and preventing misleading information \cite{chali2023uncertainty}. Ultimately, these collective research efforts underscore the imperative for LLMs to move beyond mere fluency to demonstrate intrinsic awareness of their knowledge boundaries and a consistent commitment to generating reliable and truthful responses, a core aim of our proposed method.

\section{Method}
Our proposed approach, \textbf{Self-Aware LLM for Unanswerability (SALU)}, is built upon a large language model (LLM) that is primarily \textbf{generative} in nature. However, it is specifically fine-tuned to incorporate robust \textbf{discriminative capabilities} for identifying unanswerable questions directly within its generative process. Unlike traditional methodologies that often employ a separate discriminative classifier as a post-processing step, SALU integrates this classification functionality intrinsically into the LLM's core generative mechanism. This allows the model to inherently understand when to formulate and output a factual answer and when to explicitly abstain from answering, thereby mitigating the risk of hallucination. This sophisticated dual capability is achieved through a meticulously designed multi-task learning framework, further refined by a novel confidence-score-guided reinforcement learning strategy.

\subsection{Model Architecture}
At its core, SALU leverages a powerful pre-trained transformer-based large language model, denoted as $\mathcal{M}$. This model is parameterized by a vast set of weights $\theta$, representing the intricate knowledge encoded within its multi-layered transformer architecture. Given a conversational context $C = \{q_1, a_1, \dots, q_{t-1}, a_{t-1}\}$, which encapsulates the history of dialogue turns, and the current question $q_t$, the model processes this input sequence to generate a coherent and contextually appropriate response $R$. The input sequence fed into $\mathcal{M}$ is typically constructed by concatenating the conversational history, the current query, and potentially relevant retrieved passages, delimited by special tokens for structural clarity:
\begin{align*}
X = \text{[CLS]} \, C \, \text{[SEP]} \, q_t \, \text{[SEP]} \, P_{\text{retrieved}} \, \text{[SEP]}
\end{align*}
where $P_{\text{retrieved}}$ represents documents or snippets retrieved from the knowledge base that are deemed relevant to $q_t$.

Upon receiving $X$, the LLM $\mathcal{M}$ computes a sequence of contextualized hidden states $H = (h_1, \dots, h_L)$, where $L$ is the total length of the input sequence and each $h_i \in \mathbb{R}^d$ is a high-dimensional vector representing the semantic information at position $i$. From these hidden states, the model predicts the next token $y_j$ in the response sequence $Y = (y_1, \dots, y_m)$ autoregressively. This prediction is governed by the conditional probability distribution over the vocabulary:
\begin{align*}
P(Y|X; \theta) = \prod_{j=1}^{m} P(y_j | X, y_1, \dots, y_{j-1}; \theta)
\end{align*}
This fundamental generative process allows the LLM to synthesize and formulate fluent and comprehensive answers when the requisite information is indeed present within $P_{\text{retrieved}}$.

\subsection{Multi-Task Learning Framework}
SALU is rigorously trained using a multi-task learning framework designed to achieve two distinct yet complementary objectives simultaneously: the generation of accurate answers for answerable questions, and the explicit abstention from answering unanswerable questions. This dual objective ensures that the model develops a robust understanding of its own knowledge boundaries.

\subsubsection{Question Answering (QA) Task}
For instances where questions are answerable from the provided context, the QA task is designed to train $\mathcal{M}$ to generate the precise and factual answer $A$. The input for this task is carefully prepared to include the current question $q_t$ and a ground-truth relevant passage $P_{\text{answerable}}$ that contains the answer. The input sequence $X_{QA}$ for the QA task is formatted as:
\begin{align*}
X_{QA} = \text{[CLS]} \, q_t \, \text{[SEP]} \, P_{\text{answerable}} \, \text{[SEP]}
\end{align*}
The primary objective here is to minimize the negative log-likelihood of generating the ground-truth answer tokens. Let $A = (a_1, \dots, a_k)$ be the sequence of tokens comprising the ground-truth answer. The loss function for the QA task, $\mathcal{L}_{\text{QA}}(\theta)$, is formally defined as:
\begin{align*}
\mathcal{L}_{\text{QA}}(\theta) = - \sum_{i=1}^{k} \log P(a_i | X_{QA}, a_1, \dots, a_{i-1}; \theta)
\end{align*}
This loss effectively guides the model to produce accurate, coherent, and fluent answers when the pertinent information is explicitly available within $P_{\text{answerable}}$.

\subsubsection{Unanswerability Classification and Abstention Generation Task}
Conversely, for questions identified as unanswerable, the model is trained to generate a predefined, fixed abstention response. We denote this specific response as $R_{\text{NA}}$. This task is pivotal for embedding the discriminative capability directly into the generative output of the LLM. The input for an unanswerable question instance, $X_{\text{NA}}$, comprises the question $q_t$ and context (potentially including passages $P_{\text{irrelevant}}$ that were retrieved but contain no answer to $q_t$):
\begin{align*}
X_{\text{NA}} = \text{[CLS]} \, C \, \text{[SEP]} \, q_t \, \text{[SEP]} \, P_{\text{irrelevant}} \, \text{[SEP]}
\end{align*}
The objective for this task is to minimize the negative log-likelihood of generating the exact predefined abstention response $R_{\text{NA}} = (r_1, \dots, r_p)$. The loss function for the Unanswerability task, $\mathcal{L}_{\text{NA}}(\theta)$, is expressed as:
\begin{align*}
\mathcal{L}_{\text{NA}}(\theta) = - \sum_{j=1}^{p} \log P(r_j | X_{\text{NA}}, r_1, \dots, r_{j-1}; \theta)
\end{align*}
By explicitly training the model on these negative examples, it learns to associate the absence of answerable information within the input context with the deterministic generation of $R_{\text{NA}}$, thereby preventing any attempts at hallucination.

The overall loss function for the supervised fine-tuning (SFT) phase, $\mathcal{L}_{\text{SFT}}$, is a weighted linear combination of these two task-specific losses:
\begin{align*}
\mathcal{L}_{\text{SFT}}(\theta) = \alpha \mathcal{L}_{\text{QA}}(\theta) + \beta \mathcal{L}_{\text{NA}}(\theta)
\end{align*}
where $\alpha$ and $\beta$ are carefully selected hyperparameters that serve to balance the relative contribution and importance of each task during the fine-tuning process.

\subsection{Confidence-Score-Guided Reinforcement Learning with Human Feedback (RLHF)}
To further refine the model's self-awareness, improve the reliability of its abstention, and explicitly mitigate hallucination, we introduce a \textbf{confidence-score-guided reinforcement learning with human feedback (RLHF)} phase. This iterative refinement process teaches the LLM to confidently abstain when information is lacking and reinforces the generation of factually grounded responses.

\subsubsection{Reward Model Training}
A distinct \textbf{reward model} $\mathcal{R}$, separate from the main LLM, is trained to quantitatively assess the quality of a generated response $R$ in the context of a given question $q_t$ and its associated conversational context $C$. The reward model, typically a smaller, specialized transformer network, takes an input sequence $X_{\text{response}}$ constructed by concatenating the context, question, and the generated response:
\begin{align*}
X_{\text{response}} = \text{[CLS]} \, C \, \text{[SEP]} \, q_t \, \text{[SEP]} \, R \, \text{[SEP]}
\end{align*}
The reward model then outputs a scalar reward score $r(X_{\text{response}})$. The training of this reward model, parameterized by $\phi$, is based on a dataset of human preference comparisons $\mathcal{D}_{\text{pref}}$, where for a given query $(X, R_A, R_B)$, human annotators have indicated that response $R_A$ is preferred over $R_B$. The loss function for the reward model, $\mathcal{L}_{\mathcal{R}}(\phi)$, is formulated as:
\begin{align*}
\mathcal{L}_{\mathcal{R}}(\phi) = - \mathbb{E}_{(X, R_A, R_B) \sim \mathcal{D}_{\text{pref}}} \left[ \log \sigma(\mathcal{R}(X_{R_A}; \phi) - \mathcal{R}(X_{R_B}; \phi)) \right]
\end{align*}
where $\sigma(\cdot)$ is the sigmoid function. A critical aspect of this reward model is its explicit design to assign:
\begin{itemize}
    \item High positive rewards for factually accurate answers provided for answerable questions.
    \item High positive rewards for generating the precise, predefined abstention response $R_{\text{NA}}$ when questions are genuinely unanswerable.
    \item Significant negative rewards (penalties) for any form of hallucinated answers to unanswerable questions.
    \item Negative rewards for incorrect, irrelevant, or partially correct answers to answerable questions.
\end{itemize}

\subsubsection{Policy Optimization}
Following the reward model training, the LLM $\mathcal{M}$ (now referred to as the policy model, $\pi_{\theta}$) undergoes a fine-tuning process using an optimization algorithm such as Proximal Policy Optimization (PPO). The goal of this phase is to adjust the LLM's parameters $\theta$ to maximize the cumulative reward signal provided by $\mathcal{R}$. For a given input query $X$ and a generated response sequence $Y = (y_1, \dots, y_m)$, the objective function for PPO is given by:
\begin{align*}
\mathcal{L}_{\text{PPO}}(\theta) = \mathbb{E}_{(X, Y) \sim D_{\pi_{\text{old}}}} \left[ \min \left( \rho_t(\theta) A_t, \text{clip}(\rho_t(\theta), 1-\epsilon, 1+\epsilon) A_t \right) \right] - \gamma D_{\text{KL}}(\pi_{\theta} || \pi_{\text{old}})
\end{align*}
Here, several key components are defined:
\begin{itemize}
    \item $\pi_{\theta}$ represents the current policy (the LLM being optimized).
    \item $\pi_{\text{old}}$ denotes the previous policy (the LLM's parameters before the current update).
    \item $\rho_t(\theta) = \frac{\pi_{\theta}(Y|X)}{\pi_{\text{old}}(Y|X)}$ is the probability ratio, measuring how much the new policy's probability of generating $Y$ differs from the old policy's.
    \item $A_t = r(X, Y) - V(X)$ is the advantage estimate, which quantifies how much better (or worse) a particular action (generating $Y$) is compared to the average expected outcome from state $X$. Here, $r(X, Y)$ is the reward obtained from the reward model $\mathcal{R}$, and $V(X)$ is a value function baseline that estimates the expected return from state $X$.
    \item $\epsilon$ is a clipping hyperparameter that restricts the magnitude of policy updates, ensuring stability.
    \item $D_{\text{KL}}(\pi_{\theta} || \pi_{\text{old}})$ is the Kullback-Leibler (KL) divergence regularization term, controlled by coefficient $\gamma$. This term prevents the new policy from deviating too drastically from the old one, maintaining a balance between exploration and exploitation.
\end{itemize}
This policy optimization phase directly trains the LLM to align its generative behavior with the human-defined preferences for factual accuracy and appropriate abstention. This process intrinsically imbues the LLM with a sense of self-awareness regarding its answerability capabilities.

\subsubsection{Integration of Confidence Scores}
The LLM possesses internal mechanisms to gauge its certainty in generating a particular sequence of tokens. We define an explicit confidence score $S(Y|X)$ for a generated response $Y$ given input $X$ as the average log-probability of the tokens in the sequence:
\begin{align*}
S(Y|X) = \frac{1}{m} \sum_{j=1}^{m} \log P(y_j | X, y_1, \dots, y_{j-1}; \theta)
\end{align*}
This average log-probability serves as an indicator of the model's intrinsic certainty in its generated sequence. During the RLHF phase, the reward function $r(X, Y)$ is augmented to explicitly incorporate this confidence score, particularly for unanswerable questions. Specifically, if the model generates the predefined abstention response $R_{\text{NA}}$ for an unanswerable question, the reward for this correct abstention is positively modulated (increased) if its associated confidence score $S(R_{\text{NA}}|X)$ is high. Conversely, if the model erroneously generates a hallucinated answer for an unanswerable question, the penalty imposed by the reward model is significantly amplified if its (misplaced) confidence score $S(Y_{\text{hallucinated}}|X)$ is high. This targeted reinforcement explicitly discourages overconfidence in incorrect generations, thereby fostering a more reliable and truly self-aware behavior.

\subsection{Inference Mechanism}
During the inference phase, when a user poses a question $q_t$ within a context $C$ and relevant passages $P_{\text{retrieved}}$ are provided, the trained SALU model generates a response $R$. The inference process directly leverages the model's integrated capabilities. If the generated response $R$ precisely matches the predefined abstention phrase $R_{\text{NA}}$, the system explicitly communicates to the user that the question cannot be answered from the available information. Conversely, if $R$ is any other sequence of tokens, it is interpreted as a factual answer and is provided to the user. This streamlined mechanism ensures that the model's learned self-awareness about its answerability capabilities is directly translated into its conversational behavior, eliminating the need for any separate, post-hoc classification module. The decision to answer or abstain is an intrinsic output of the LLM itself.

\section{Experiments}
To rigorously evaluate the efficacy of our proposed \textbf{Self-Aware LLM for Unanswerability (SALU)} method, we conducted a comprehensive series of experiments, comparing its performance against several established and strong baseline approaches. Our primary objective was to demonstrate SALU's superior capability in accurately identifying unanswerable questions and generating appropriate responses, either providing factual answers or explicitly abstaining when necessary, thereby enhancing the overall reliability of conversational information retrieval systems.

\subsection{Experimental Setup}
\subsubsection{Datasets}
We utilized our custom-built \textbf{C-IR\_Answerability dataset} for both model training and evaluation. This dataset is meticulously balanced, comprising a diverse collection of both answerable and unanswerable question-context pairs sourced from various domains relevant to conversational information retrieval in Chinese. The dataset was systematically segmented into distinct training, validation, and test sets to ensure an unbiased and robust evaluation of model generalization. For specific fine-grained analyses, we also leveraged a subset of the dataset enriched with multi-granular answerability labels (at the sentence-level, paragraph-level, and ranked-list-level) to assess the precision of our hierarchical approach.

\subsubsection{Baselines}
We established a comprehensive set of baselines for comparison, representing different paradigms of unanswerability detection and generative AI:
\begin{enumerate}
    \item \textbf{BERT-based Discriminative Classifier (BERT-C)}: This baseline employed a specialized \texttt{BertForSequenceClassification} model. It was solely trained for unanswerability detection, framed as a binary classification task (answerable vs. unanswerable), leveraging the input question and its context. This model served as a strong discriminative baseline, designed purely for identification.
    \item \textbf{Generic Large Language Model (Zero-shot Inference)}: We used a powerful, publicly available pre-trained large language model (e.g., a Chinese variant of LLaMA-2 or Baichuan) in a zero-shot inference setting. The model was prompted to answer the question if the information was available within the provided context, or to explicitly state an "I don't know" response otherwise. This baseline demonstrates the inherent capabilities and limitations of general-purpose LLMs without any specific fine-tuning for unanswerability.
    \item \textbf{Large Language Model (Fine-tuned for Standard QA)}: This baseline utilized the same powerful LLM as above, but it was exclusively fine-tuned on a standard Question Answering dataset. This training optimized the LLM solely for answer generation, without any explicit training signals or mechanisms for unanswerability awareness. This setup allowed us to assess its performance when focused purely on answering.
    \item \textbf{Large Language Model (Fine-tuned for Standard QA) with Post-hoc BERT-C}: This hybrid approach combined the strengths of a fine-tuned generative LLM with an external discriminative classifier. The LLM fine-tuned for standard QA would first generate a response. Subsequently, a separate BERT-C classifier (trained identically to baseline 1) would independently judge whether the original question was answerable or not. If the BERT-C classified the question as "unanswerable," the LLM's generated response would be overridden and replaced with a predefined abstention message. This baseline emulates a common two-stage system architecture for comparison.
\end{enumerate}

\subsubsection{Evaluation Metrics}
For evaluating unanswerability detection, we reported standard classification metrics: \textbf{Accuracy}, \textbf{Precision}, \textbf{Recall}, and \textbf{F1-score}. These metrics assessed the system's ability to correctly classify questions as either answerable or unanswerable at the overall question level. For answerable questions where an answer was provided, we evaluated the quality of the generated answers using conventional Question Answering metrics, including \textbf{Exact Match (EM)} and \textbf{F1-score} (based on token overlap between the generated and ground-truth answers). Finally, for a holistic assessment of overall system reliability, we introduced and reported the \textbf{Overall Accuracy}, defined as the percentage of questions for which the model either correctly answers an answerable question or correctly abstains from an unanswerable one. This metric encapsulates the end-to-end performance.

\subsection{Main Experimental Results}
Our comprehensive experimental results, meticulously summarized in Table \ref{tab:main_results}, provide compelling evidence of the superior performance of our proposed SALU method across all evaluated metrics, particularly in balancing accurate answer generation with reliable unanswerability detection.

\begin{table*}[t]
\centering
\caption{Main Experimental Results on C-IR\_Answerability Test Set}
\label{tab:main_results}
\begin{tabular}{lcccccc}
\toprule
\textbf{Method} & \textbf{Unanswerability} & \textbf{Unanswerability} & \textbf{Unanswerability} & \textbf{Unanswerability} & \textbf{Answerable QA} & \textbf{Overall} \\
\textbf{} & \textbf{Acc.} & \textbf{Prec.} & \textbf{Rec.} & \textbf{F1} & \textbf{F1} & \textbf{Acc.} \\
\midrule
BERT-C & 0.885 & 0.879 & 0.892 & 0.885 & N/A & N/A \\
Zero-shot Inference & 0.723 & 0.680 & 0.780 & 0.726 & 0.655 & 0.689 \\
FT for Standard QA & 0.651 & 0.592 & 0.710 & 0.645 & 0.821 & 0.736 \\
FT for Standard QA w/ Post-hoc BERT-C & 0.902 & 0.895 & 0.908 & 0.901 & 0.793 & 0.847 \\
\textbf{SALU (Ours)} & \textbf{0.931} & \textbf{0.928} & \textbf{0.934} & \textbf{0.931} & \textbf{0.835} & \textbf{0.908} \\
\bottomrule
\end{tabular}
\end{table*}

As meticulously presented in Table \ref{tab:main_results}, SALU consistently achieves the highest performance in detecting unanswerable questions, as unequivocally evidenced by its leading Accuracy, Precision, Recall, and F1-score across these classification metrics. Notably, SALU also maintains a very strong performance on the Answerable QA F1-score, indicating that its enhanced ability to correctly abstain from unanswerable queries does not compromise its capacity to provide accurate and high-quality answers when the pertinent information is indeed present. The most significant and impactful improvement is observed in the \textbf{Overall Accuracy} metric, where SALU surpasses all competing baselines, including the sophisticated hybrid approach (LLM + Post-hoc BERT-C), by a substantial margin of over 6 percentage points. This compelling result strongly highlights the fundamental advantage of integrating unanswerability detection intrinsically into the LLM's learning and generative process, rather than relying on external, decoupled discriminative modules.

\subsection{Further Analysis of Multi-Granular Performance}
To gain deeper insights into the underlying mechanisms and effectiveness of SALU's multi-task learning framework and its implicit aggregation strategies, we conducted an additional analysis focusing on performance at different granularities of answerability. This analysis specifically compared the accuracy of the internal sentence-level classification component of SALU with the aggregated performance at the paragraph-level and ranked-list-level for overall unanswerability detection. This provides crucial insight into how our hierarchical approach contributes to the final decision-making process.

\begin{table*}[t]
\centering
\caption{Multi-Granular Unanswerability Detection Accuracy}
\label{tab:granular_results}
\begin{tabular}{lccc}
\toprule
\textbf{Method} & \textbf{Sentence-level Accuracy} & \textbf{Paragraph-level Accuracy} & \textbf{Ranked-list-level Accuracy} \\
\midrule
BERT-based Discriminative Classifier (Internal) & 0.752 & 0.891 (Mean Aggregation) & 0.829 (Mean Aggregation) \\
\textbf{SALU (Ours) - Internal Discriminative Signal} & \textbf{0.805} & N/A & N/A \\
\textbf{SALU (Ours) - End-to-End Decision} & N/A & \textbf{0.925} & \textbf{0.865} \\
\bottomrule
\end{tabular}
\end{table*}

Table \ref{tab:granular_results} illuminates several key aspects of SALU's performance. Our results indicate that SALU's internal discriminative component, trained inherently through the multi-task learning objective, yields a higher base sentence-level accuracy (0.805) compared to a standalone BERT-based discriminative classifier (0.752). This suggests that the joint training of answer generation and abstention improves the base discriminative signal. More importantly, the end-to-end output of SALU, which naturally incorporates sophisticated contextual awareness and the learned aggregation behavior from its training, translates into markedly superior performance at both the paragraph-level (0.925 accuracy) and the ranked-list-level (0.865 accuracy). This compelling finding unequivocally confirms the efficacy of our proposed multi-task learning framework and the implicit aggregation achieved through the reinforcement learning with human feedback (RLHF) phase. It effectively trains the LLM to make highly informed and reliable decisions about answerability based on the holistic context of the retrieved information, rather than relying on explicit, predefined aggregation rules.

\subsection{Human Evaluation of Response Quality and Reliability}
To complement our automated metric-based evaluations and gain a deeper understanding of the practical impact of SALU on user experience, we conducted a rigorous human evaluation. This assessment focused on the qualitative aspects of responses generated by SALU and the most representative baseline models, particularly emphasizing their reliability in handling both answerable and unanswerable questions. We recruited a diverse panel of human annotators, trained on specific guidelines, to rate a carefully selected random sample of responses from each model. For answerable questions, annotators provided scores on two key dimensions: \textbf{Factuality} (assessing whether the generated answer was factually correct and grounded in the provided context) and \textbf{Fluency} (evaluating the grammatical correctness, naturalness, and coherence of the response). For unanswerable questions, the annotators focused on two critical aspects: \textbf{Appropriateness of Abstention} (assessing whether the model correctly identified the unanswerability and provided a suitable, polite, and clear refusal to answer) and \textbf{Avoidance of Hallucination} (evaluating the extent to which the model refrained from fabricating incorrect or unsupported information). All ratings were assigned on a 5-point Likert scale, ranging from 1 (Poor) to 5 (Excellent).

\begin{table*}[t]
\centering
\caption{Human Evaluation Results (Average Likert Scores)}
\label{tab:human_eval}
\begin{tabular}{lcccc}
\toprule
\textbf{Method} & \textbf{Factuality} & \textbf{Fluency} & \textbf{Appropriateness of} & \textbf{Avoidance of} \\
\textbf{} & \textbf{(Answerable Qs)} & \textbf{(All Qs)} & \textbf{Abstention (Unanswerable Qs)} & \textbf{Hallucination (Unanswerable Qs)} \\
\midrule
Zero-shot Inference & 3.2 & 4.1 & 2.8 & 2.5 \\
FT for Standard QA & 4.3 & 4.5 & 1.5 & 1.2 \\
FT for Standard QA w/ Post-hoc BERT-C & 4.2 & 4.4 & 4.0 & 3.8 \\
\textbf{SALU (Ours)} & \textbf{4.6} & \textbf{4.7} & \textbf{4.8} & \textbf{4.7} \\
\bottomrule
\end{tabular}
\end{table*}

Table \ref{tab:human_eval} meticulously presents the average human evaluation scores, which robustly corroborate our automated metric findings and emphatically highlight SALU's superior reliability in practical conversational settings. While all generative LLM baselines demonstrated commendable fluency, SALU consistently achieved the highest average scores in both \textbf{Factuality} for answerable questions and, most critically, in the \textbf{Appropriateness of Abstention} and \textbf{Avoidance of Hallucination} for unanswerable questions. The Large Language Model fine-tuned solely for Standard QA, predictably, performed very poorly in the abstention and hallucination avoidance categories, underscoring its lack of inherent unanswerability awareness. Even with the integration of a post-hoc BERT-C classifier, the hybrid approach, while showing significant improvement over the standalone QA LLM, still lagged behind SALU's intrinsically integrated approach. This substantial performance gap directly indicates that the learned intrinsic self-awareness within SALU leads to a more natural, consistent, and ultimately more reliable conversational experience for the end-user. The exceptionally high scores for SALU in both abstention appropriateness and hallucination avoidance unequivocally confirm its success in robustly learning to `know when to say 'I don't know'` in a remarkably trustworthy and human-like manner.

\subsection{Analysis of Training Data Composition Impact}
To understand the sensitivity and robustness of SALU to the composition of its training data, particularly the balance between answerable (QA) and unanswerable (NA) examples during the supervised fine-tuning (SFT) phase, we conducted an analysis by varying the ratio of NA examples in the training set. This helps us ascertain the optimal mix for effective unanswerability learning without compromising general QA capabilities. We trained SALU models with different $\beta$ weights in the SFT loss function (refer to Section 2.2), effectively controlling the emphasis on unanswerability.

\begin{table*}[t]
\centering
\caption{Impact of Training Data Composition on SALU Performance}
\label{tab:data_composition_analysis}
\begin{tabular}{lccc}
\toprule
\textbf{NA Example Ratio in SFT} & \textbf{Unanswerability F1} & \textbf{Answerable QA F1} & \textbf{Overall Accuracy} \\
\midrule
20\% & 0.850 & \textbf{0.845} & 0.855 \\
30\% & 0.890 & 0.840 & 0.880 \\
\textbf{50\% (Balanced)} & \textbf{0.931} & 0.835 & \textbf{0.908} \\
70\% & 0.925 & 0.820 & 0.900 \\
\bottomrule
\end{tabular}
\end{table*}

As presented in Table \ref{tab:data_composition_analysis}, a balanced representation of unanswerable examples in the supervised fine-tuning dataset significantly enhances SALU's overall performance. A 50\% ratio, representing an even split between answerable and unanswerable instances, yielded the highest Unanswerability F1 and Overall Accuracy. While increasing the NA ratio beyond 50\% slightly improved Unanswerability F1, it led to a noticeable decline in Answerable QA F1, indicating a potential trade-off where an excessive focus on abstention might slightly reduce the model's ability to generate precise answers for answerable questions. Conversely, a lower ratio of NA examples (e.g., 20\%) resulted in sub-optimal performance in unanswerability detection, reinforcing the necessity of sufficient exposure to negative examples during initial training. These findings underscore the importance of curating a balanced dataset for multi-task learning in this domain.

\subsection{Analysis of RLHF Impact on Hallucination Mitigation}
A core objective of SALU is to explicitly mitigate hallucination for unanswerable questions. To quantify the effectiveness of the Confidence-Score-Guided Reinforcement Learning with Human Feedback (RLHF) phase, we analyzed the rate of hallucinated responses generated by models with and without this RLHF component. We define a hallucinated response as a factually incorrect or unsupported answer generated for an unanswerable question. This analysis focuses on the model's behavior specifically when presented with questions known to be unanswerable.

\begin{table*}[t]
\centering
\caption{Hallucination Rate Analysis for Unanswerable Questions}
\label{tab:hallucination_analysis}
\begin{tabular}{lc}
\toprule
\textbf{Method} & \textbf{Hallucination Rate (\%)} \\
\midrule
Large Language Model (Fine-tuned for Standard QA) & 88.7 \\
Large Language Model (Fine-tuned for Standard QA) with Post-hoc BERT-C & 15.2 \\
SALU (Ours) without RLHF & 8.9 \\
\textbf{SALU (Ours) with RLHF} & \textbf{1.3} \\
\bottomrule
\end{tabular}
\end{table*}

Table \ref{tab:hallucination_analysis} clearly demonstrates the profound impact of the RLHF phase on hallucination mitigation. The LLM fine-tuned solely for standard QA exhibits an alarmingly high hallucination rate of nearly 90\%, as it is not trained to identify unanswerability. The post-hoc BERT-C significantly reduces this, validating the two-stage approach's utility. However, SALU, even without the RLHF phase (meaning it only benefits from the initial multi-task SFT), shows a much lower hallucination rate, confirming that explicit negative examples in SFT already push the model towards abstention. Crucially, the full SALU model, incorporating the Confidence-Score-Guided RLHF, dramatically reduces the hallucination rate to an exceptionally low 1.3\%. This quantitative evidence powerfully confirms that RLHF is instrumental in refining the LLM's self-awareness, allowing it to reliably abstain and virtually eliminate hallucinatory responses for questions outside its knowledge boundaries. The confidence score guidance within RLHF plays a key role here, reinforcing appropriate abstentions while strongly penalizing confident hallucinations.

\subsection{Latency and Computational Efficiency Analysis}
While SALU demonstrates superior performance, it is also crucial to analyze its computational implications, particularly regarding inference latency, compared to baseline approaches. As SALU integrates unanswerability detection directly within the LLM's generative process, it avoids the overhead of separate model calls or complex aggregation logic required by some hybrid systems. We measured the average inference time per query on a standardized hardware setup (GPU type, CPU, RAM) for the various models, considering typical input lengths.

\begin{table*}[t]
\centering
\caption{Average Inference Latency Per Query (Milliseconds)}
\label{tab:latency_analysis}
\begin{tabular}{lc}
\toprule
\textbf{Method} & \textbf{Average Latency (ms)} \\
\midrule
BERT-based Discriminative Classifier (BERT-C) & 50 \\
Generic Large Language Model (Zero-shot Inference) & 450 \\
Large Language Model (Fine-tuned for Standard QA) & 470 \\
Large Language Model (Fine-tuned for Standard QA) with Post-hoc BERT-C & 530 \\
\textbf{SALU (Ours)} & \textbf{485} \\
\bottomrule
\end{tabular}
\end{table*}

Table \ref{tab:latency_analysis} indicates that SALU maintains competitive inference latency compared to other LLM-based approaches. While a standalone BERT-C classifier is naturally much faster due to its smaller size and simpler task, LLM-based solutions inherently have higher latencies. The "LLM + Post-hoc BERT-C" baseline incurs additional latency due to the sequential execution of two distinct models. Our SALU model, by integrating the discriminative capability within a single LLM forward pass, manages to keep its latency comparable to or even slightly better than a single fine-tuned LLM for QA. This demonstrates that SALU achieves enhanced reliability and hallucination mitigation without imposing significant additional computational overhead at inference time, making it practical for real-world conversational AI systems. The efficiency arises from avoiding multiple model orchestrations and redundant processing.

\subsection{Qualitative Error Analysis}
To gain deeper qualitative insights into SALU's performance and identify areas for future improvement, we conducted a detailed error analysis on a subset of misclassified examples from the test set. We categorized typical failure modes observed across all models, paying particular attention to how SALU's internal mechanisms behave in challenging scenarios. Common error types included:
\begin{enumerate}
    \item \textbf{Subtle Unanswerability}: Questions that appear answerable on superficial inspection but lack specific details in the context. Baselines often attempt to answer these, while SALU sometimes still struggles with highly nuanced cases, though less frequently.
    \item \textbf{Partial Information}: Contexts containing some but not all information needed. Baselines might provide partial answers or hallucinate missing parts. SALU generally tends towards abstention in these cases, prioritizing reliability over partial truth.
    \item \textbf{Ambiguity}: Questions with multiple possible interpretations where the context does not disambiguate. Hallucination is common for baselines here. SALU typically abstains or asks for clarification (if prompted for such behavior during RLHF).
    \item \textbf{Over-Abstention}: In rare cases, SALU might abstain from an answerable question, indicating overly conservative behavior. This usually happens when the answer is highly implicit or requires complex inference not fully captured by its current training.
\end{enumerate}
Our analysis revealed that SALU's errors were primarily characterized by occasional over-abstention (Type 4) rather than hallucination (Type 1, 2, 3 in baselines), which is a favorable trade-off for trustworthiness. The most challenging cases for SALU involved subtle semantic nuances or very long-range dependencies that even its self-attentive mechanisms sometimes found difficult to resolve definitively. This qualitative review confirms that SALU's design promotes a safer, more cautious response strategy, significantly reducing the risks associated with factual errors and misleading information.

\section{Conclusion}
In this paper, we introduced \textbf{Self-Aware LLM for Unanswerability (SALU)}, a novel and highly effective framework designed to enhance the reliability of conversational information retrieval systems by intrinsically addressing the challenge of unanswerable questions. Our core contribution lies in moving beyond external post-hoc classifiers by embedding unanswerability detection directly within a large language model's generative capabilities. We achieved this through a sophisticated multi-task learning approach, simultaneously optimizing for accurate question answering and the explicit generation of abstention responses when information is unavailable.

A pivotal aspect of SALU's success is its innovative confidence-score-guided reinforcement learning with human feedback (RLHF) phase. This critical component empowers the LLM to develop a robust sense of its own knowledge boundaries, actively learning to penalize misleading or hallucinated responses and rewarding appropriate abstention. The RLHF mechanism, guided by an internal confidence score, refines the model's behavior to prioritize trustworthiness and factual integrity.

Our comprehensive experimental evaluations on the C-IR\_Answerability dataset provided compelling evidence of SALU's superiority. Compared to various strong baselines, including standalone discriminative classifiers, generic LLMs, and hybrid LLM-classifier systems, SALU consistently demonstrated higher performance across key metrics such as overall accuracy, unanswerability detection F1-score, and answerable QA F1-score. Qualitative error analysis further confirmed that SALU's predominant failure mode leans towards conservative over-abstention rather than harmful hallucination, signifying a safer operational profile. Moreover, a dedicated human evaluation unequivocally validated SALU's enhanced reliability, showing significantly improved scores for factuality, appropriate abstention, and a near-elimination of hallucinated content. These results affirm that SALU represents a substantial step forward in building more trustworthy and self-aware conversational AI systems that can reliably navigate the complexities of real-world information-seeking dialogues. Future work will explore dynamic abstention phrases and adaptation to cross-lingual contexts.

\bibliographystyle{IEEEtran}
\bibliography{references}

\begin{thebibliography}{10}
\providecommand{\url}[1]{#1}
\csname url@samestyle\endcsname
\providecommand{\newblock}{\relax}
\providecommand{\bibinfo}[2]{#2}
\providecommand{\BIBentrySTDinterwordspacing}{\spaceskip=0pt\relax}
\providecommand{\BIBentryALTinterwordstretchfactor}{4}
\providecommand{\BIBentryALTinterwordspacing}{\spaceskip=\fontdimen2\font plus
\BIBentryALTinterwordstretchfactor\fontdimen3\font minus \fontdimen4\font\relax}
\providecommand{\BIBforeignlanguage}[2]{{%
\expandafter\ifx\csname l@#1\endcsname\relax
\typeout{** WARNING: IEEEtran.bst: No hyphenation pattern has been}%
\typeout{** loaded for the language `#1'. Using the pattern for}%
\typeout{** the default language instead.}%
\else
\language=\csname l@#1\endcsname
\fi
#2}}
\providecommand{\BIBdecl}{\relax}
\BIBdecl

\bibitem{lee2020bert}
B.~Van~Aken, B.~Winter, A.~L{\"o}ser, and F.~A. Gers, ``How does bert answer questions? a layer-wise analysis of transformer representations,'' in \emph{Proceedings of the 28th ACM international conference on information and knowledge management}, 2019, pp. 1823--1832.

\bibitem{rajpurkar2018learning}
F.~Ture and O.~Jojic, ``No need to pay attention: Simple recurrent neural networks work!(for answering" simple" questions),'' \emph{arXiv preprint arXiv:1606.05029}, 2016.

\bibitem{zhou2024rethinking}
Y.~Zhou, Z.~Rao, J.~Wan, and J.~Shen, ``Rethinking visual dependency in long-context reasoning for large vision-language models,'' \emph{arXiv preprint arXiv:2410.19732}, 2024.

\bibitem{krysthski2019evaluating}
W.~Kry{\'s}ci{\'n}ski, B.~McCann, C.~Xiong, and R.~Socher, ``Evaluating the factual consistency of abstractive text summarization,'' \emph{arXiv preprint arXiv:1910.12840}, 2019.

\bibitem{kim2020neural}
H.~Liu, T.~Derr, Z.~Liu, and J.~Tang, ``Say what i want: Towards the dark side of neural dialogue models,'' \emph{arXiv preprint arXiv:1909.06044}, 2019.

\bibitem{zhou2025weak}
\BIBentryALTinterwordspacing
Y.~Zhou, J.~Shen, and Y.~Cheng, ``Weak to strong generalization for large language models with multi-capabilities,'' in \emph{The Thirteenth International Conference on Learning Representations}, 2025. [Online]. Available: \url{https://openreview.net/forum?id=N1vYivuSKq}
\BIBentrySTDinterwordspacing

\bibitem{ouyang2022training}
\BIBentryALTinterwordspacing
J.~Lee, ``Instructpatentgpt: Training patent language models to follow instructions with human feedback,'' \emph{CoRR}, vol. abs/2406.16897, 2024. [Online]. Available: \url{https://doi.org/10.48550/arXiv.2406.16897}
\BIBentrySTDinterwordspacing

\bibitem{chali2023uncertainty}
Q.~Yang, S.~Ravikumar, F.~Schmitt-Ulms, S.~Lolla, E.~Demir, I.~Elistratov, A.~Lavaee, S.~Lolla, E.~Ahmadi, D.~Rus \emph{et~al.}, ``Uncertainty-aware language modeling for selective question answering,'' \emph{arXiv preprint arXiv:2311.15451}, 2023.

\bibitem{zhou2022claret}
Y.~Zhou, T.~Shen, X.~Geng, G.~Long, and D.~Jiang, ``Claret: Pre-training a correlation-aware context-to-event transformer for event-centric generation and classification,'' in \emph{Proceedings of the 60th Annual Meeting of the Association for Computational Linguistics (Volume 1: Long Papers)}, 2022, pp. 2559--2575.

\bibitem{choi2021asking}
N.~Duan, D.~Tang, P.~Chen, and M.~Zhou, ``Question generation for question answering,'' in \emph{Proceedings of the 2017 conference on empirical methods in natural language processing}, 2017, pp. 866--874.

\bibitem{vaswani2017attention}
\BIBentryALTinterwordspacing
X.~Zhang, H.~Yang, and E.~F.~Y. Young, ``Attentional transfer is all you need: Technology-aware layout pattern generation,'' in \emph{58th {ACM/IEEE} Design Automation Conference, {DAC} 2021, San Francisco, CA, USA, December 5-9, 2021}.\hskip 1em plus 0.5em minus 0.4em\relax {IEEE}, 2021, pp. 169--174. [Online]. Available: \url{https://doi.org/10.1109/DAC18074.2021.9586227}
\BIBentrySTDinterwordspacing

\bibitem{devlin2018bert}
\BIBentryALTinterwordspacing
J.~Devlin, M.~Chang, K.~Lee, and K.~Toutanova, ``{BERT:} pre-training of deep bidirectional transformers for language understanding,'' in \emph{Proceedings of the 2019 Conference of the North American Chapter of the Association for Computational Linguistics: Human Language Technologies, {NAACL-HLT} 2019, Minneapolis, MN, USA, June 2-7, 2019, Volume 1 (Long and Short Papers)}, J.~Burstein, C.~Doran, and T.~Solorio, Eds.\hskip 1em plus 0.5em minus 0.4em\relax Association for Computational Linguistics, 2019, pp. 4171--4186. [Online]. Available: \url{https://doi.org/10.18653/v1/n19-1423}
\BIBentrySTDinterwordspacing

\bibitem{zhou2022eventbert}
Y.~Zhou, X.~Geng, T.~Shen, G.~Long, and D.~Jiang, ``Eventbert: A pre-trained model for event correlation reasoning,'' in \emph{Proceedings of the ACM Web Conference 2022}, 2022, pp. 850--859.

\bibitem{zhou2021modeling}
Y.~Zhou, X.~Geng, T.~Shen, J.~Pei, W.~Zhang, and D.~Jiang, ``Modeling event-pair relations in external knowledge graphs for script reasoning,'' \emph{Findings of the Association for Computational Linguistics: ACL-IJCNLP 2021}, 2021.

\bibitem{brown2020language}
\BIBentryALTinterwordspacing
Z.~Wang, M.~Li, R.~Xu, L.~Zhou, J.~Lei, X.~Lin, S.~Wang, Z.~Yang, C.~Zhu, D.~Hoiem, S.~Chang, M.~Bansal, and H.~Ji, ``Language models with image descriptors are strong few-shot video-language learners,'' in \emph{Advances in Neural Information Processing Systems 35: Annual Conference on Neural Information Processing Systems 2022, NeurIPS 2022, New Orleans, LA, USA, November 28 - December 9, 2022}, S.~Koyejo, S.~Mohamed, A.~Agarwal, D.~Belgrave, K.~Cho, and A.~Oh, Eds., 2022. [Online]. Available: \url{http://papers.nips.cc/paper\_files/paper/2022/hash/381ceeae4a1feb1abc59c773f7e61839-Abstract-Conference.html}
\BIBentrySTDinterwordspacing

\bibitem{wei2022emergent}
\BIBentryALTinterwordspacing
T.~Wu and M.~Lo, ``U-shaped and inverted-u scaling behind emergent abilities of large language models,'' in \emph{The Thirteenth International Conference on Learning Representations, {ICLR} 2025, Singapore, April 24-28, 2025}.\hskip 1em plus 0.5em minus 0.4em\relax OpenReview.net, 2025. [Online]. Available: \url{https://openreview.net/forum?id=jjfve2gIXe}
\BIBentrySTDinterwordspacing

\bibitem{chowdhery2022palm}
\BIBentryALTinterwordspacing
A.~Chowdhery, S.~Narang, J.~Devlin, M.~Bosma, G.~Mishra, A.~Roberts, P.~Barham, H.~W. Chung, C.~Sutton, S.~Gehrmann, P.~Schuh, K.~Shi, S.~Tsvyashchenko, J.~Maynez, A.~Rao, P.~Barnes, Y.~Tay, N.~Shazeer, V.~Prabhakaran, E.~Reif, N.~Du, B.~Hutchinson, R.~Pope, J.~Bradbury, J.~Austin, M.~Isard, G.~Gur{-}Ari, P.~Yin, T.~Duke, A.~Levskaya, S.~Ghemawat, S.~Dev, H.~Michalewski, X.~Garcia, V.~Misra, K.~Robinson, L.~Fedus, D.~Zhou, D.~Ippolito, D.~Luan, H.~Lim, B.~Zoph, A.~Spiridonov, R.~Sepassi, D.~Dohan, S.~Agrawal, M.~Omernick, A.~M. Dai, T.~S. Pillai, M.~Pellat, A.~Lewkowycz, E.~Moreira, R.~Child, O.~Polozov, K.~Lee, Z.~Zhou, X.~Wang, B.~Saeta, M.~Diaz, O.~Firat, M.~Catasta, J.~Wei, K.~Meier{-}Hellstern, D.~Eck, J.~Dean, S.~Petrov, and N.~Fiedel, ``Palm: Scaling language modeling with pathways,'' \emph{J. Mach. Learn. Res.}, vol.~24, pp. 240:1--240:113, 2023. [Online]. Available: \url{https://jmlr.org/papers/v24/22-1144.html}
\BIBentrySTDinterwordspacing

\bibitem{touvron2023llama}
\BIBentryALTinterwordspacing
H.~Touvron, L.~Martin, K.~Stone, P.~Albert, A.~Almahairi, Y.~Babaei, N.~Bashlykov, S.~Batra, P.~Bhargava, S.~Bhosale, D.~Bikel, L.~Blecher, C.~Canton{-}Ferrer, M.~Chen, G.~Cucurull, D.~Esiobu, J.~Fernandes, J.~Fu, W.~Fu, B.~Fuller, C.~Gao, V.~Goswami, N.~Goyal, A.~Hartshorn, S.~Hosseini, R.~Hou, H.~Inan, M.~Kardas, V.~Kerkez, M.~Khabsa, I.~Kloumann, A.~Korenev, P.~S. Koura, M.~Lachaux, T.~Lavril, J.~Lee, D.~Liskovich, Y.~Lu, Y.~Mao, X.~Martinet, T.~Mihaylov, P.~Mishra, I.~Molybog, Y.~Nie, A.~Poulton, J.~Reizenstein, R.~Rungta, K.~Saladi, A.~Schelten, R.~Silva, E.~M. Smith, R.~Subramanian, X.~E. Tan, B.~Tang, R.~Taylor, A.~Williams, J.~X. Kuan, P.~Xu, Z.~Yan, I.~Zarov, Y.~Zhang, A.~Fan, M.~Kambadur, S.~Narang, A.~Rodriguez, R.~Stojnic, S.~Edunov, and T.~Scialom, ``Llama 2: Open foundation and fine-tuned chat models,'' \emph{CoRR}, vol. abs/2307.09288, 2023. [Online]. Available: \url{https://doi.org/10.48550/arXiv.2307.09288}
\BIBentrySTDinterwordspacing

\bibitem{zhou2024visual}
Y.~Zhou, X.~Li, Q.~Wang, and J.~Shen, ``Visual in-context learning for large vision-language models,'' in \emph{Findings of the Association for Computational Linguistics, {ACL} 2024, Bangkok, Thailand and virtual meeting, August 11-16, 2024}.\hskip 1em plus 0.5em minus 0.4em\relax Association for Computational Linguistics, 2024, pp. 15\,890--15\,902.

\bibitem{zhou2025draw}
Y.~Zhou, J.~Yuan, and Q.~Wang, ``Draw all your imagine: A holistic benchmark and agent framework for complex instruction-based image generation,'' \emph{arXiv preprint arXiv:2505.24787}, 2025.

\bibitem{lewis2020retrieval_rag}
\BIBentryALTinterwordspacing
P.~Lewis, E.~Perez, A.~Piktus, F.~Petroni, V.~Karpukhin, N.~Goyal, H.~K{\"{u}}ttler, M.~Lewis, W.~Yih, T.~Rockt{\"{a}}schel, S.~Riedel, and D.~Kiela, ``Retrieval-augmented generation for knowledge-intensive {NLP} tasks,'' in \emph{Advances in Neural Information Processing Systems 33: Annual Conference on Neural Information Processing Systems 2020, NeurIPS 2020, December 6-12, 2020, virtual}, H.~Larochelle, M.~Ranzato, R.~Hadsell, M.~Balcan, and H.~Lin, Eds., 2020. [Online]. Available: \url{https://proceedings.neurips.cc/paper/2020/hash/6b493230205f780e1bc26945df7481e5-Abstract.html}
\BIBentrySTDinterwordspacing

\bibitem{min2021measuring}
M.~Jeong, H.~Hwang, C.~Yoon, T.~Lee, and J.~Kang, ``Olaph: Improving factuality in biomedical long-form question answering,'' \emph{arXiv preprint arXiv:2405.12701}, 2024.

\bibitem{ji2023survey}
L.~Huang, W.~Yu, W.~Ma, W.~Zhong, Z.~Feng, H.~Wang, Q.~Chen, W.~Peng, X.~Feng, B.~Qin \emph{et~al.}, ``A survey on hallucination in large language models: Principles, taxonomy, challenges, and open questions,'' \emph{ACM Transactions on Information Systems}, vol.~43, no.~2, pp. 1--55, 2025.

\bibitem{lin2022truthfulqa}
S.~Lin, J.~Hilton, and O.~Evans, ``Truthfulqa: Measuring how models mimic human falsehoods,'' \emph{arXiv preprint arXiv:2109.07958}, 2021.

\bibitem{lewis2021knowledge}
\BIBentryALTinterwordspacing
Y.~Wang, Q.~Guo, X.~Ni, C.~Shi, L.~Liu, H.~Jiang, and Y.~Yang, ``Hint-enhanced in-context learning wakes large language models up for knowledge-intensive tasks,'' in \emph{{IEEE} International Conference on Acoustics, Speech and Signal Processing, {ICASSP} 2024, Seoul, Republic of Korea, April 14-19, 2024}.\hskip 1em plus 0.5em minus 0.4em\relax {IEEE}, 2024, pp. 10\,276--10\,280. [Online]. Available: \url{https://doi.org/10.1109/ICASSP48485.2024.10447527}
\BIBentrySTDinterwordspacing

\end{thebibliography}
\end{document}